\documentclass[letterpaper]{article}

\usepackage{ijcai17}
\usepackage{times}
\usepackage{helvet}
\usepackage{courier}

\usepackage{hyperref}
\usepackage{fancyhdr}
\usepackage{datetime}
\usdate

\newcommand{\NumberOfQuestions}{244}
\newtheorem{task}{Task Definition}

\begin{document}
\title{Using Thought-Provoking Children's Questions \\
to Drive Artificial Intelligence Research}
\author{Erik T. Mueller \and Henry Minsky \\
Minsky Institute for Artificial Intelligence \\
\url{http://minskyinstitute.org/}}
\maketitle
\begin{center}
\today\ \currenttime
\end{center}
\begin{small}
\begin{abstract}
We propose to use thought-provoking children's questions (TPCQs), namely \emph{Highlights\/} BrainPlay questions, as a new method to drive artificial intelligence research and to evaluate the capabilities of general-purpose AI systems.
These questions are designed to stimulate thought and learning in children,
and they can be used to do the same thing in AI systems, while demonstrating the system's reasoning capabilities to the evaluator.
We introduce the TPCQ task, which which takes a TPCQ question
as input and produces as output (1) answers to the question and
(2) learned generalizations.
We discuss how BrainPlay questions stimulate learning.
We analyze \NumberOfQuestions{} BrainPlay questions, and we report
statistics on question type, question class, answer cardinality,
answer class, types of knowledge needed, and types of reasoning
needed.
We find that BrainPlay questions span many aspects of intelligence.
Because the answers to BrainPlay questions and the generalizations
learned from them are often highly open-ended, we suggest using human
judges for evaluation.
\end{abstract}
\end{small}

\section{Introduction}

As artificial intelligence tasks like
fact-based question answering \cite{FerrucciEtAl:2012}
and face recognition \cite{TaigmanYangRanzatoWolf:2014}
become mostly solved, there is a need for harder tasks.
Consider the following questions from the children's magazine
\emph{Highlights\/}:
\begin{quote}
Why doesn't every key open every lock?

Which is older, a tree or a leaf on the tree?

Why aren't pants pockets as big as backpacks?

Flags wave, people wave, and the ocean has waves. How are these waves alike?

What part of a fish is farthest from the head?

Is an ice-cream cone wider at the bottom or at the top?

Could you sing a song in a dark room? Could you put together a puzzle? 

Why can't you move faster than your shadow?

What might happen if you put a bee in your pocket?

If you could not remember today's date, what are five ways you could find out?
\end{quote}
Although these questions are short and designed to be answered by young children, they are very hard for computers.
It is unlikely that the reader has heard these questions before, and yet correct answers can be produced by most children within seconds, as well as explanations of the reasoning behind these answers.
The embarrassing fact is that answering and learning from these questions
is way beyond the capabilities of existing AI systems.
They are wide open.

We propose answering and learning from
\emph{thought-provoking children's questions\/} (TPCQs), which are
available in the BrainPlay\footnote{BrainPlay questions are published each month in \emph{Highlights\/}, which is available from \url{https://www.highlights.com/}.}
column of \emph{Highlights\/}, as a useful metric
for driving research and evaluating general-purpose AI systems.
TPCQs test a system's ability to make novel connections, which is necessary
for intelligence.

While this method does require that the system have a powerful language facility, this is a crucial capability for a large class of useful AI systems. Without the system having the capacity to understand and generate language, it is extremely difficult for researchers to communicate abstract goals and tasks to the system, to draw its attention to salient topics, to receive answers to questions, and for the system to explain its behavior.
\begin{task}[TPCQ]
\emph{Given a thought-provoking children's question $Q$, produce
\begin{itemize}
\item
one or more answers to the question $A_{1}, A_{2}, \ldots{}$
\item
one or more learned generalizations $L_{1}, L_{2}, \ldots{}$
\end{itemize}}
\end{task}
Example: \\
$Q$: Name three animals that hatch from eggs. \\
$A_{1}$: birds \\
$A_{2}$: chickens \\
$A_{3}$: ducks \\
$A_{4}$: snakes \\
$L_{1}$: Animals with feathers hatch from eggs.

Learning doesn't only happen through real experiences and doesn't
always require the addition of new knowledge.
A hallmark of human-level intelligence is the ability to combine existing knowledge through \emph{imagined\/} situations, to answer questions which you have never before been asked. You may have pieces of knowledge whose connections are not apparent
until someone pushes you to notice them.
This is what BrainPlay questions are designed to do.

\section{Highlights BrainPlay}

\emph{Highlights\/} magazine was started in 1946 by Garry Cleveland
Myers and Caroline Clark Myers.
The magazine includes a BrainPlay column, which was called Headwork before
November 2004. In this paper, we use the term BrainPlay for both Headwork
and BrainPlay questions.

The \emph{Highlights\/} editors develop BrainPlay questions with great care.
The questions are designed to ``[stimulate] children from five to twelve to
think and reason by working over in their heads what is already there,
arriving at new ideas not learned from books''
\cite{Myers:1968}.

For example, consider the BrainPlay question ``In a room with a staircase leading to the second floor, how can you figure out the height of the first-floor ceiling?''
This question suggests a novel technique:
to measure the height of a ceiling when there is a staircase
leading up to the next floor, multiply the rise of the steps by the
number of steps.

BrainPlay first appeared in the second issue of \emph{Highlights\/}
in September 1946 \cite{Wood:1986}.
Each month, BrainPlay presents around 20 questions arranged
by age level \cite{MyersMyers:1964}.
Correct answers to the questions aren't provided.

\section{Analysis of BrainPlay Questions}

To get an idea of what we're up against, we performed an analysis of
BrainPlay questions in the \emph{Highlights\/} issues from January 2000
to December 2000.
We started by segmenting each top-level question into one or more
subquestions.
For example, the top-level question
\begin{quote}
Would you rather wear a hood or a hat? Why?
\end{quote}
is segmented into a first question and a second question.

Table~\ref{table:subquestion-composition} shows the composition
of subquestions.
\begin{table}
\begin{center}
\begin{tabular}{|r|r|r|} \hline
\textbf{Subquestion} & \textbf{\% Questions} & \textbf{\# Questions}\\ \hline
1st & 65.07\% & 244/375\\
2nd & 22.13\% & 83/375\\
3rd & 6.40\% & 24/375\\
4th & 3.47\% & 13/375\\
5th & 1.60\% & 6/375\\
6th & 0.80\% & 3/375\\
7th & 0.27\% & 1/375\\
8th & 0.27\% & 1/375\\
\hline
\end{tabular}
\end{center}
\caption{Subquestion Composition\label{table:subquestion-composition}}
\end{table}
Table~\ref{table:subquestion-length} gives statistics on the
length of subquestions.
The first question tends to be the longest.
\begin{table}
\begin{center}
\begin{tabular}{|r|r|r|r|r|} \hline
\textbf{Subquestion} & \textbf{Mean} & \textbf{Min} & \textbf{Max} & \textbf{SD}\\ \hline
1st & 11.59 & 3 & 34 & 5.95\\
2nd & 4.10 & 1 & 16 & 2.81\\
3rd & 3.58 & 1 & 10 & 2.68\\
4th & 3.69 & 1 & 10 & 3.02\\
5th & 3.50 & 1 & 7 & 2.29\\
6th & 4.33 & 1 & 7 & 2.49\\
7th & 5.00 & 5 & 5 & 0.00\\
8th & 5.00 & 5 & 5 & 0.00\\
\hline
\end{tabular}
\end{center}
\caption{Subquestion Length (number of words, SD = standard deviation)\label{table:subquestion-length}}
\end{table}
The second and following questions typically
ask for explanations for the answer to the first question,
ask variations on the first question (often involving coreference),
or follow up in some other way.
For the remainder of the analysis, we considered only first questions.

We annotated each first question with exactly one
\emph{question type\/},
\emph{question class\/},
\emph{answer cardinality\/}, and
\emph{answer class\/}, and we annotated each first question with one or more
\emph{types of knowledge needed\/}
and
\emph{types of reasoning needed\/}.
We developed an initial set of annotation tags like
\textbf{Open-Ended} and \textbf{What-If}
and revised them as needed during the annotation process.

\subsection{Question Type}

Statistics on the question type are shown in Table~\ref{table:question-type}.
\begin{table}
\begin{center}
\begin{tabular}{|l|r|r|} \hline
\textbf{Type} & \textbf{\% Questions} & \textbf{\# Questions}\\ \hline
Open-Ended & 87.30\% & 213/244\\
Multiple Choice & 11.48\% & 28/244\\
Yes-No & 1.23\% & 3/244\\
\hline
\end{tabular}
\end{center}
\caption{Question Type\label{table:question-type}}
\end{table}
\begin{small}
\begin{description}
\item[Open-Ended] \emph{Answer choices aren't provided.\/} \\
What is your favorite way to travel? \\
Name three uses for bells.
\item[Multiple Choice] \emph{Answer choices are provided, and the question
is not a Yes-No question.\/} \\
Would you rather receive a phone call or a letter? \\
Is it harder to ride a bike or to run fast?
\item[Yes-No] \emph{The answer choices are \emph{yes\/} and \emph{no\/}.\/} \\
Do you know anyone else with your initials? \\
Have you ever cried because you were very happy?
\end{description}
\end{small}

\subsection{Question Class}

Statistics on the question class are shown in Table~\ref{table:question-class}.
\begin{table}
\begin{center}
\begin{tabular}{|l|r|r|} \hline
\textbf{Class} & \textbf{\% Questions} & \textbf{\# Questions}\\ \hline
Facts & 14.34\% & 35/244\\
Caring & 10.66\% & 26/244\\
What-If & 9.02\% & 22/244\\
Comparative & 8.20\% & 20/244\\
Personal Experience & 6.97\% & 17/244\\
Personal Preference & 6.56\% & 16/244\\
Theory of Mind & 6.15\% & 15/244\\
Purpose & 5.74\% & 14/244\\
Difference & 5.33\% & 13/244\\
Reason & 4.92\% & 12/244\\
Meaning & 4.51\% & 11/244\\
Action & 4.10\% & 10/244\\
Personal Facts & 3.28\% & 8/244\\
Similarity & 2.87\% & 7/244\\
Superlative & 2.46\% & 6/244\\
Debugging & 2.05\% & 5/244\\
Description & 2.05\% & 5/244\\
Count & 0.41\% & 1/244\\
Sort & 0.41\% & 1/244\\
\hline
\end{tabular}
\end{center}
\caption{Question Class\label{table:question-class}}
\end{table}
\begin{small}
\begin{description}
\item[Facts] \emph{ Asks about facts (may require reasoning).\/}
\\Name three animals that hatch from eggs.
\\How does a turtle protect itself?
\item[Caring] \emph{ Stimulates thought about caring and kindness.\/}
\\What could you do today to help someone else?
\\If your family has company, what can you do to be a good host?
\item[What-If] \emph{ Asks about a hypothetical scenario.\/}
\\If you had a pet that could talk, what would the two of you talk about?
\\If you could change your schedule at school, what would you change?
\item[Comparative] \emph{ Involves a comparative.\/}
\\Is it easier to swallow a pill or a spoonful of medicine?
\\Would it be easier to remember the date of a party or the date of a haircut appointment?
\item[Personal Experience] \emph{ Asks about personal experiences.\/}
\\Have you ever been so busy that you forgot to eat a meal?
\\What popular sayings did you first hear in a song or movie?
\item[Personal Preference] \emph{ Asks about personal preferences.\/}
\\Describe your favorite place to go for a walk.
\\If you could meet any person in the world, who would it be?
\item[Theory of Mind] \emph{ Evaluates theory of mind \cite{Doherty:2009}.\/}
\\Ryan looked at the sliced apple and said, ``This must have been sliced a while ago.'' How might he have known?
\\When Otis arrived at the pool, he quickly figured out which person was the new swim coach. How might he have guessed?
\item[Purpose] \emph{ Asks about the purpose or function of something.\/}
\\What tools do you need for drawing?
\\Name three uses for bells.
\item[Difference] \emph{ Asks for the differences between two things.\/}
\\How is taking a music lesson different from playing music on your own?
\\What's the difference between a riddle and a joke?
\item[Reason] \emph{ Asks about the reason for something.\/}
\\Why do babies cry more often than adults?
\\Why do we frame paintings and photos before hanging them up?
\item[Meaning] \emph{ Asks for the meaning of a word or phrase.\/}
\\What is meant by the saying ``Money doesn't grow on trees''?
\\What does it mean to ``go the extra mile''?
\item[Action] \emph{ Asks for an action to be performed like singing or drawing.\/}
\\Draw a heart in the air with your finger.
\\Make a hand signal that means ``good job.''
\item[Personal Facts] \emph{ Asks about personal facts.\/}
\\Are you ticklish?
\\How many teeth do you have?
\item[Similarity] \emph{ Asks for the similarities between two things.\/}
\\How are socks and mittens alike?
\\How is honey like maple syrup?
\item[Superlative] \emph{ Involves a superlative.\/}
\\What is the best smell in spring?
\\Where do you laugh the most: at school, at home, or with friends?
\item[Debugging] \emph{ Requires debugging of a problem or situation.\/}
\\When Erik looked at his plane tickets, it seemed as if his flight from Oregon to Rhode Island would take six hours longer than his flight from Rhode Island to Oregon. Why was this?
\\Jackson and his family were watching TV when suddenly they lost reception. What might have caused this?
\item[Description] \emph{ Asks for a description.\/}
\\Describe some rocks you've seen.
\\Describe how a wheel works.
\item[Count] \emph{ Asks for a count.\/}
\\How many pets do you know by name?
\item[Sort] \emph{ Asks for items to be sorted by some attribute.\/}
\\List these in order of size: moon, bird, star, airplane.
\end{description}
\end{small}

A number of questions involve personal experiences, preferences, and facts.
The answers to these questions are person-dependent.
How shall we deal with these?
The first reaction might be simply to throw them out.
But consider that an intelligent, autonomous AI system will have its own
personal experiences and preferences.
These are essential aspects of a general-purpose AI system.
Therefore it would be a mistake to throw these questions out.
Because there is no gold standard answer key for them,
answers can be judged for plausibility by human judges,
as in the Turing test \cite{Turing:1950}.

Some questions request an action to be performed.
Again, we could throw these out, but then we would be throwing out
some of the most revealing questions.
Instead, the system can perform the actions in a three-dimensional
simulator (or in the world if the system has a body),
and the results can be judged by humans.

Human judging is more time-consuming, but it is currently the best way
of evaluating novel, previously unseen answers to novel, previously unseen
questions.

A question like ``Have you ever been so busy that you forgot to eat
a meal?'' makes sense for an AI system, because the question probes
essential knowledge of goals, plans, and mental states.
General-purpose AI systems must be able to recognize, remember, and apply
concepts like ``being busy'' and ``forgetting to perform a task.''

\subsection{Answer Cardinality}

Statistics on how many answers are required by a question are shown in
Table~\ref{table:answer-cardinality}.
\begin{table}
\begin{center}
\begin{tabular}{|l|r|r|} \hline
\textbf{Cardinality} & \textbf{\% Questions} & \textbf{\# Questions}\\ \hline
1 & 49.59\% & 121/244\\
$>$1 & 45.90\% & 112/244\\
3 & 3.28\% & 8/244\\
2 & 0.82\% & 2/244\\
5 & 0.41\% & 1/244\\
\hline
\end{tabular}
\end{center}
\caption{Answer Cardinality\label{table:answer-cardinality}}
\end{table}
\begin{small}
\begin{description}
\item[1] \emph{ One answer.\/}
\\Who is the tallest person you know?
\\Is it easier to throw or to catch a ball?
\item[$>$1] \emph{ More than one answer.\/}
\\How are a bird's wings different from a butterfly's wings?
\\Why do people make New Year's resolutions?
\item[2] \emph{ Two answers.\/}
\\What weather and location are ideal for stargazing?
\\Think of a fruit and a vegetable that begin with the letter p.
\item[3] \emph{ Three answers.\/}
\\Name three ways to have fun on a rainy day.
\\Name three objects that are shaped like a triangle.
\item[5] \emph{ Five answers.\/}
\\List the top five things that you like to do with your friends.
\end{description}
\end{small}

\subsection{Answer Class}

Statistics on the answer class are shown in Table~\ref{table:answer-class}.
A gold standard answer key can be developed for questions of class
\textbf{Exactly One} and \textbf{Several}.
Thus the answers to 103 (42.21\%) of the 244 BrainPlay questions
we analyzed can be evaluated automatically.

What about the remaining questions?
Human judging will be needed for the answers to questions of class
\textbf{Many}, \textbf{Personal},
\textbf{Open}, \textbf{Debatable}, and
\textbf{Nontextual Answer}.
More points should be awarded for correct answers to harder questions.

\begin{table}
\begin{center}
\begin{tabular}{|l|r|r|} \hline
\textbf{Class} & \textbf{\% Questions} & \textbf{\# Questions}\\ \hline
Many & 24.18\% & 59/244\\
Exactly One & 22.13\% & 54/244\\
Several & 20.08\% & 49/244\\
Personal & 18.85\% & 46/244\\
Open & 9.02\% & 22/244\\
Debatable & 3.69\% & 9/244\\
Nontextual Answer & 2.05\% & 5/244\\
\hline
\end{tabular}
\end{center}
\caption{Answer Class\label{table:answer-class}}
\end{table}
\begin{small}
\begin{description}
\item[Many] \emph{ The question has many short, correct answers.\/}
\\When might it be useful to know some jokes?
\\Where can you find spiders?
\item[Exactly One] \emph{ The question has a single possible correct answer.\/}
\\During which season do you usually wear sunglasses?
\\What does it mean to be ``on cloud nine''?
\item[Several] \emph{ The question has a few short, correct answers.\/}
\\What kinds of things do you write about in a diary?
\\Name three animals that hatch from eggs.
\item[Personal] \emph{ The question can only be answered relative to personal experience.\/}
\\Try to name all of the people you have talked with today.
\\Would you rather receive a phone call or a letter?
\item[Open] \emph{ The question has many possibly long answers.\/}
\\What might happen if televisions everywhere stopped working?
\\If you had a pet that could talk, what would the two of you talk about?
\item[Debatable] \emph{ It is difficult to judge the correctness of the answer.\/}
\\Is it easier to swallow a pill or a spoonful of medicine?
\\Is it harder to ride a bike or to run fast?
\item[Nontextual Answer] \emph{ The question cannot be answered using text. Instead, it requires an action to be performed.\/}
\\Try to clap your hands behind your back.
\\Sing part of a song you know.
\end{description}
\end{small}

\subsection{Types of Knowledge Needed}

Statistics on the types of knowledge needed to answer questions
are shown in Table~\ref{table:knowledge-needed}.
The percentages sum to more than 100 because each question is annotated
with one or more types of knowledge.

\begin{table}
\begin{center}
\begin{tabular}{|l|r|r|} \hline
\textbf{Knowledge Type} & \textbf{\% Questions} & \textbf{\# Questions}\\ \hline
Scripts & 29.51\% & 72/244\\
Plans/Goals & 15.16\% & 37/244\\
Physics & 11.89\% & 29/244\\
Properties/Attributes & 11.48\% & 28/244\\
Human Body & 11.48\% & 28/244\\
Relations & 11.07\% & 27/244\\
Interpersonal Relations & 11.07\% & 27/244\\
Episodic Memory & 9.84\% & 24/244\\
Devices/Appliances & 9.02\% & 22/244\\
Mental States & 7.38\% & 18/244\\
Animals & 6.56\% & 16/244\\
Lexicon & 6.15\% & 15/244\\
Emotions & 4.92\% & 12/244\\
Shapes & 3.28\% & 8/244\\
Sounds & 2.87\% & 7/244\\
Location & 2.46\% & 6/244\\
Plants & 2.46\% & 6/244\\
Food & 2.46\% & 6/244\\
Weather & 2.46\% & 6/244\\
Letters & 1.23\% & 3/244\\
Taste & 0.82\% & 2/244\\
Smell & 0.82\% & 2/244\\
\hline
\end{tabular}
\end{center}
\caption{Knowledge Needed\label{table:knowledge-needed}}
\end{table}
\begin{small}
\begin{description}
\item[Scripts] \emph{ Stereotypical situations and scripts \cite{SchankAbelson:1977}.\/}
\\Name a game you can play alone.
\\Would you rather receive a phone call or a letter?
\item[Plans/Goals] \emph{ Goals and plans \cite{SchankAbelson:1977}.\/}
\\Why do people make New Year's resolutions?
\\What are the benefits of working on a project with others?
\item[Physics] \emph{ Physics.\/}
\\Is it easier to throw or to catch a ball?
\\Try to clap your hands behind your back.
\item[Properties/Attributes] \emph{ Properties and attributes of people and things.\/}
\\What kinds of hats are casual?
\\How are a snake and an eel similar?
\item[Human Body] \emph{ The human body.\/}
\\Try to make your body into the shape of each letter in your name.
\\What do elbows and knees have in common?
\item[Relations] \emph{ Database relations involving people or things.\/}
\\Whose phone numbers do you know by heart?
\\Which is higher, clouds or the sun?
\item[Interpersonal Relations] \emph{ Interpersonal relations \cite{Heider:1958}.\/}
\\If a friend lied to you, how could he or she regain your trust?
\\List the top five things that you like to do with your friends.
\item[Episodic Memory] \emph{ Episodic memory \cite{Tulving:1983,Hasselmo:2012}.\/}
\\Try to name all of the people you have talked with today.
\\Tell about a time when you felt proud of someone.
\item[Devices/Appliances] \emph{ Devices.\/}
\\What tools do you need for drawing?
\\What is let in or kept out by windows?
\item[Mental States] \emph{ Mental states.\/}
\\Why do people make New Year's resolutions?
\\If a friend lied to you, how could he or she regain your trust?
\item[Animals] \emph{ Animals.\/}
\\Why might a bear with a cub be more dangerous than a bear by itself?
\\Name an animal that can walk as soon as it is born.
\item[Lexicon] \emph{ English lexicon or dictionary.\/}
\\What does it mean to be ``on cloud nine''?
\\What does it mean to ``go the extra mile''?
\item[Emotions] \emph{ Human emotions.\/}
\\Describe how it feels to watch someone opening a gift that you gave.
\\How can you tell when someone is nervous about something?
\item[Shapes] \emph{ Shapes of objects.\/}
\\Name three objects that are shaped like a triangle.
\\Draw polka dots.
\item[Sounds] \emph{ Sounds.\/}
\\What noise would a dragon make?
\\What kinds of shoes are noisy?
\item[Location] \emph{ Locations and places.\/}
\\Describe your favorite place to go for a walk.
\\Name three jobs that involve working outdoors.
\item[Plants] \emph{ Plants.\/}
\\During which season might you rake leaves?
\\What makes a salad a salad?
\item[Food] \emph{ Food and cooking.\/}
\\Think of a fruit and a vegetable that begin with the letter p.
\\Name three foods that are purple.
\item[Weather] \emph{ Weather.\/}
\\Name three ways to have fun on a rainy day.
\\Where is the safest place to be during a thunderstorm?
\item[Letters] \emph{ The alphabet and letters.\/}
\\Try to make your body into the shape of each letter in your name.
\\Which letters of the alphabet can you draw using only curved lines?
\item[Taste] \emph{ Taste.\/}
\\Name three foods that might cause you to make a face when you eat them.
\item[Smell] \emph{ Smell.\/}
\\What is the best smell in spring?
\end{description}
\end{small}

\subsection{Types of Reasoning Needed}

Statistics on the types of reasoning needed to answer questions
are shown in Table~\ref{table:reasoning-needed}.
Again, the percentages sum to more than 100 because each question is
annotated with one or more reasoning types.
\begin{table}
\begin{center}
\begin{tabular}{|l|r|r|} \hline
\textbf{Reasoning Type} & \textbf{\% Questions} & \textbf{\# Questions}\\ \hline
Database Retrieval & 37.70\% & 92/244\\
Simulation & 24.59\% & 60/244\\
Planning & 22.54\% & 55/244\\
Comparison & 18.85\% & 46/244\\
Episodic Memory & 9.84\% & 24/244\\
Visualization & 8.61\% & 21/244\\
3D Simulation & 7.79\% & 19/244\\
Invention & 3.28\% & 8/244\\
Arithmetic & 1.23\% & 3/244\\
\hline
\end{tabular}
\end{center}
\caption{Reasoning Needed\label{table:reasoning-needed}}
\end{table}
\begin{small}
\begin{description}
\item[Database Retrieval] \emph{ Database retrieval.\/}
\\Who is the tallest person you know?
\\Name three animals that hatch from eggs.
\item[Simulation] \emph{ Simulation of the course of events, not necessarily requiring physical or three-dimensional reasoning.\/}
\\Is it easier to swallow a pill or a spoonful of medicine?
\\Why might a bear with a cub be more dangerous than a bear by itself?
\item[Planning] \emph{ Planning or generating a sequence of actions to achieve a goal \cite{GhallabNauTraverso:2004}.\/}
\\What might happen if televisions everywhere stopped working?
\\Describe your favorite place to go for a walk.
\item[Comparison] \emph{ Quantitative or qualitative comparison.\/}
\\Who is the tallest person you know?
\\What do elbows and knees have in common?
\item[Episodic Memory] \emph{ Retrieving or recalling personal experiences from episodic memory.\/}
\\Have you ever been so busy that you forgot to eat a meal?
\\What mistakes have you made that you've learned from?
\item[Visualization] \emph{ Visualization and imagery.\/}
\\How are a bird's wings different from a butterfly's wings?
\\Of the stars, the moon, and the sun, which can be seen during the day?
\item[3D Simulation] \emph{ Physical or three-dimensional simulation.\/}
\\Try to clap your hands behind your back.
\\Why don't we wear watches on our ankles?
\item[Invention] \emph{ Inventing or creating something.\/}
\\Describe a toy that you would like to invent.
\\Make up a word that means ``so funny you can't stop laughing.''
\item[Arithmetic] \emph{ Arithmetic operations.\/}
\\How many inches have you grown in the past year?
\\In what year will you be able to register to vote?
\end{description}
\end{small}

\subsection{Correlation with Question Position}

The correlation of various annotations with position in the
BrainPlay column is given in
Table~\ref{table:positional-correlation}.
Only correlations with magnitude above 0.1 are shown.
The \emph{Highlights\/} editors present the BrainPlay questions in increasing
order of difficulty \cite{MyersMyers:1964},
so these correlations give a rough idea of difficulty.
High positive correlations correspond to high difficulty, whereas
high negative correlations correspond to low difficulty.
\begin{table}
\begin{center}
\begin{tabular}{|l|r|} \hline
\textbf{Tag} & \textbf{Correlation}\\ \hline
Caring & 0.3059\\
Planning & 0.2599\\
Plans/Goals & 0.2293\\
Interpersonal Relations & 0.1838\\
Scripts & 0.1396\\
Simulation & 0.1269\\
Arithmetic & 0.1263\\
5 & 0.1017\\
Properties/Attributes & -0.1005\\
Personal Experience & -0.1081\\
Location & -0.1083\\
Plants & -0.1308\\
Letters & -0.1330\\
Database Retrieval & -0.1413\\
Nontextual Answer & -0.1741\\
Human Body & -0.1923\\
Action & -0.2242\\
\hline
\end{tabular}
\end{center}
\caption{Correlation with Question Position\label{table:positional-correlation}}
\end{table}

\section{BrainPlay's Coverage of Intelligence}

We can use the major sections of the fifth edition of
\emph{The Cognitive Neurosciences\/} \cite{GazzanigaMangun:2014}
as a guide to the many areas of human intelligence.
A rough correspondence between these sections and BrainPlay is shown in
Table~\ref{table:Gazzaniga-Mangun-coverage}.
(``VI Memory'' includes prediction and imagination.)
\begin{table}
\begin{center}
\begin{tabular}{|l|l|} \hline
{\bf Gazzaniga/Mangun Part}  & {\bf BrainPlay} \\ \hline
I Developmental and & \\
Evolutionary Cognitive & \\
Neuroscience & \\
\hline
II Plasticity and Learning & learning from questions \\
\hline
III Visual Attention   & \\
\hline
IV Sensation and Perception & Shapes, Sounds, Smell \\
                            & Visualization \\
\hline
V Motor Systems and Action & Action, Planning \\
\hline
VI Memory     & Episodic Memory, Facts \\
              & Scripts, What-If \\
              & Simulation \\
\hline
VII Language and Abstract & Meaning, Lexicon \\
Thought & Description \\
\hline
VIII Social Neuroscience & Emotions, Caring \\
and Emotion         & Interpersonal Relations \\
                    & Theory of Mind \\
\hline
IX Consciousness & \\
\hline
X Advances in Methodology & \\
\hline
XI Neuroscience and Society & \\
\hline
\end{tabular}
\end{center}
\caption{Correspondence of Gazzaniga and Mangun (2014) sections and BrainPlay\label{table:Gazzaniga-Mangun-coverage}}
\end{table}
We see that BrainPlay questions span many aspects of intelligence.

By design and intent, many of the thought-provoking children's questions are designed to push the system into generating new knowledge, because many of the answers are open-ended and most often unlikely to have been seen before and stored explicitly. It is a hallmark of human-level intelligence that new knowledge can and often must be generated from existing knowledge when needed to accomplish a novel goal, and these questions are designed to exercise and expose those mechanisms.

\section{Related Work}

In Aristo \cite{Clark:2015}, a multiple choice elementary school science
exam question is taken as input, and an answer is produced as output.
Whereas Aristo probes science knowledge studied in school,
the BrainPlay/TPCQ task explores knowledge any child acquires
simply through experience.
Elementary science exam questions evaluate understanding of connections learned in school, while TPCQs encourage creation of new connections.

In the bAbI tasks \cite{WestonEtAl:2015}, a simple story and question
about the story are taken as input, and an answer is produced as output.
The stories are generated using a simulator based on a simple world
containing characters and objects.
The questions are very simple and restricted compared to TPCQs.

The MCTest dataset 
\cite{RichardsonBurgesRenshaw:2013} consists of short stories, multiple
choice questions about the stories, and correct answers to the questions.
The questions were designed such that answering them (1) requires information
from two or more story sentences and (2) does not require a knowledge base.
MCTest questions evaluate the ability to read, understand, and combine
information provided in a text.
TPCQs require knowledge and experience not provided in the question.

In the recognizing textual entailment (RTE) task \cite{DaganEtAl:2013},
a text $T$ and a hypothesis $H$ are taken as input, and a label
$T$ entails $H$, $H$ contradicts $T$, or \emph{unknown\/} is produced
as output.
RTE is quite general, and resources that recognize entailment could be
used as resources for performing the TPCQ task.
The Winograd schema (WS) challenge \cite{LevesqueDavisMorgenstern:2012}
is a variant of the RTE task more heavily focused on reasoning.

In the VQA task \cite{AntolEtAl:2015}, an image and a multiple choice
or open-ended question about the image are taken as input, and an answer
is produced as output.
The VQA task often involves significant reasoning, like the TPCQ task.

At the Center for Brains, Minds and Machines, the
Turing$^{++}$ questions on images \cite{PoggioMeyers:2016}
will be used to evaluate not only a system's responses to questions,
but also how accurately the system matches human behavior and neural
physiology.
The system will be compared with fMRI and MEG recordings in humans
and monkeys.

\section{Conclusion}

\emph{Highlights\/} BrainPlay questions can be answered by young children.
If today's artificial intelligence systems can't even answer these questions,
how can we really say that they are intelligent?
We believe that building systems that can answer and learn from BrainPlay
questions will increase progress in artificial intelligence.

\section*{Acknowledgments}
We thank Kent Johnson, CEO of Highlights for Children, Inc., for permission
to use the BrainPlay questions.
We also thank Patricia M. Mikelson and Sharon M. Umnik at Highlights for
providing us with the BrainPlay material.

\bibliographystyle{named}
\bibliography{tpcq}

\end{document}